\documentclass[lettersize,journal]{IEEEtran}
\usepackage{graphicx}
\usepackage{color}
\usepackage{amsmath}
\usepackage{algorithm}
\usepackage{algpseudocode}
\begin{document}

\title{An Elliptic Kernel Unsupervised Autoencoder -Graph Convolutional Network Ensemble Model for Hyperspectral Unmixing}

\author{Estefanía~Alfaro-Mejía,~\IEEEmembership{Student Member,~IEEE,}
        Carlos~J~Delgado,~\IEEEmembership{Student Member,~IEEE,}
        and~Vidya~Manian,~\IEEEmembership{Senior~Member,~IEEE}\thanks{Estefanía Alfaro-Mejía, Carlos J Delgado, and Vidya Manian are with the Artificial Intelligence Imaging Group (AIIG) Laboratory, Department of Electrical Engineering at the University of Puerto Rico, Mayaguez, PR 00681.(e-mail: estefania.alfaro@upr.edu; carlos.delgado13@upr.edu; vidya.manian@upr.edu)}\thanks{Manuscript received May XX, 2024; revised XX XX, 2024. This research is partially funded by the Artificial Intelligence Imaging Group (AIIG), Department of Electrical and Computer Engineering, University of Puerto Rico, Mayaguez. The APC is funded by NASA, grant number 80NSSC19M0155. The APC is funded by 80NSSC21M0155. (Corresponding author: Estefanía Alfaro-Mejía estefania.alfaro@upr.edu)}\thanks{Digital Object Identifier XXXX/XXX-XXX}}
        
\markboth{Journal of \LaTeX\ Class Files,~Vol.~XX, No.~XX, May~2024}%
{Alfaro \MakeLowercase{\textit{et al.}}: Bare Demo of IEEEtran.cls for Journals}

\maketitle

\begin{abstract}
Spectral Unmixing is an important technique in remote sensing used to analyze hyperspectral images to identify endmembers and estimate abundance maps. Over the past few decades, performance of techniques for endmember extraction and fractional abundance map estimation have significantly improved. This article presents an ensemble model workflow called Autoencoder Graph Ensemble Model (AEGEM) designed to extract endmembers and fractional abundance maps. An elliptical kernel is applied to measure spectral distances, generating the adjacency matrix within the elliptical neighborhood. This information is used to construct an elliptical graph, with centroids as senders and remaining pixels within the geometry as receivers. The next step involves stacking abundance maps, senders, and receivers as inputs to a Graph Convolutional Network, which processes this input to refine abundance maps. Finally, an ensemble decision-making process determines the best abundance maps based on root mean square error metric. The proposed AEGEM is assessed with benchmark datasets such as Samson, Jasper, and Urban, outperforming results obtained by baseline algorithms. For the Samson dataset, AEGEM excels in three abundance maps: water, tree and soil yielding values of $0.081$, $0.158$, and $0.182$, respectively. For the Jasper dataset, results are improved for the tree and water endmembers with values of $0.035$ and $0.060$ in that order, as well as for the mean average of the spectral angle distance metric $0.109$. For the Urban dataset, AEGEM outperforms previous results for the abundance maps of roof and asphalt, achieving values of $0.135$ and $0.240$, respectively. Additionally, for the endmembers of grass and roof, AEGEM achieves values of $0.063$ and $0.094$. 
\end{abstract}


\IEEEpeerreviewmaketitle

\section{Introduction}

\label{sec:introduction}
\IEEEPARstart{H}{yperspectral} images (HSI) offer a non-invasive approach to analyze regions of interest in an acquired scene, enabling applications in various fields, including medicine \cite{Zhou2021}-\cite{Khan2021}, food \cite{Wang2021}-\cite{Fan2021}-\cite{Benelli2021}, precision agriculture \cite{Ball2022}-\cite{Costa2021}, and remote sensing \cite{Chang2022}-\cite{Hu2022}-\cite{Liu2022}. HSI consists of material reflectances collected over multiple narrow bands within the electromagnetic spectrum, with the wavelength range varying according to the sensor used for data acquisition. This information provides enriched spectral data, allowing the analysis of materials and land cover compositions. It enables spectral unmixing (SU) analysis, which studies materials based on spectral signatures known as endmembers, and the extraction of the fractional abundance maps, in order to analyze the spectral composition that each material exhibits \cite{Borsoi2021}.

SU is an approach for determining the spectral signatures of materials, which has traditionally been addressed through classical approaches based on the linear mixing model (LMM). LMM assumes that a pixel is a linear combination of spectral signatures, weighted by their respective abundances. However, these models encounter issues related to spectral variability, a common behavior in HSI due to the low spatial resolution causing mixed pixels. 

In contrast, geometric models treat SU as a volume minimization applied to a convex surface, where the vertices represent pure materials, and the contained pixels are considered mixed pixels \cite{BioucasDias2012}. Another approach involves pure pixel algorithms, assuming the presence of at least one pure pixel for each endmember in the acquired scene. Algorithms such as pure pixel \cite{boardman1993} and N-FINDR \cite{Winter1999} are commonly employed, offering advantages in computational simplicity, although they may face challenges with mixed pixels. To address the mixed pixel problem, algorithms based on deep learning models have been proposed, with unsupervised autoencoders, including variational autoencoders \cite{Jin2023gp}, and blind convolutional autoencoder model proposed in \cite{AlfaroMeja2023} for endmember extraction without relying on previous information about sample labeling.

SU is also utilized as a preprocessing stage. In \cite{Guo2021}, a convolutional autoencoder is proposed for abundance map estimation. Furthermore, in \cite{Nalepa2021}, a deep learning ensemble algorithm is introduced for hyperspectral image classification and SU analysis. Additionally, the application of change detection based on SU is suggested using a convolutional neural network in \cite{Seydi2021}. Further, SU applications have been used for change detection to identify temporal changes in a region \cite{Wang2023}, \cite{Lin2023}, \cite{Sun2023}. In \cite{Chen2020}, an adaptive parameter estimation is proposed to address the interference of coherent speckle noise in synthetic radar images. HSI plays a crucial role in land cover analysis through classification and segmentation approaches \cite{PrezGarca2023}, \cite{Lv2023}. 

SU has been conducted using deep learning methods, including deep autoencoders \cite{Palsson2021} and blind and shallow autoencoders \cite{Palsson2022}. Typically, these autoencoders share similarities in architectural design. In the encoder, abundance maps are estimated, while the decoder performs endmembers extraction and image reconstruction. However, in \cite{Ranasinghe2020}, the proposed model conducts endmember extraction in the encoder and abundance maps estimation in the decoder stages. To effectively address SU, it is crucial to ensure the imposition of abundance nonnegativity constraint (ANC) and abundance sum-to-one constraint (ASC) to constraint abundance maps estimation and endmember extraction, respectively.  Also, the choice of loss function is critical, often relying on Mahalanobis distance, mean square error, spectral angle distance, or a combination of multiple cost functions. As SU involves both endmember extraction and fractional abundance map estimation, certain models focus on one of these aspects. For instance, \cite{Tulczyjew2022} introduces a model specifically designed for endmembers extraction.

SU has also been addressed by extracting spatial relations using graphs, In \cite{Jin2023}, a graph attention convolutional autoencoder is proposed to perform the endmembers extraction and fractional abundances map estimation. Graph convolutional networks (GCNs) have demonstrated superior performance in HSI classification by effectively combining both spatial and spectral features. In the work by Hong et al. \cite{Hong2021}, mini graph neural network approaches are introduced to tackle the challenge of computational complexity without compromising accuracy. Additionally, Jia et al \cite{Jia2024} propose a graph-in-graph approach incorporating information from the pixels and neighborhoods through segmentation techniques. 

Furthermore, SU analysis is a trending topic, commonly addressed using classical techniques \cite{Shah2020}-\cite{Drumetz2020}. However, the reconstruction of abundance maps for large spatially extended regions of interest is often inaccurate. To overcome the disadvantages of geometric approaches, unsupervised deep learning techniques have been applied achieving better results in endmembers extraction \cite{Yang2022}. Despite this, the estimation of abundance maps for large regions of interest still produces inaccurate results.

In this work, a novel initial autoencoder for endmember extraction and abundance maps estimation is developed. The proposed workflow further refines and extracts spectral features to improve abundance maps for large spatial extents, such as water, trees, and roofs.

This article introduces and assesses an ensemble model based on a convolutional autoencoder and a graph convolutional neural network, representing a significant contribution to the SU field for endmember extraction and fractional abundance maps estimation. The workflow is illustrated in Figure \ref{proposed_pipeline} and comprises eight stages. The model is called autoencoder graph ensemble model (AEGEM). The main contributions are summarized as follows:

\begin{itemize}
    \item A spatial-spectral analysis utilizing two deep learning architectures is proposed: an autoencoder, which extracts the spatial relations, and the elliptical graph, which extracts the spectral information that is input to the GCN.
    \item In order to enhance feature extraction, an elliptical kernel is introduced, which also reduces the computational effort associated with similarity measurements between pixels in the HSI.
    \item A novel adaptation of GCN for abundance maps estimation is presented.
    \item An ensemble model is proposed to fuse the best abundance maps from the autoencoder and GCN.
\end{itemize}

The manuscript is organized as follows: Section 2 briefly describes relevant concepts such as hyperspectral images, spectral unmixing, graph definition, adjacency matrix, and graph convolution neural networks. Section 3 explicitly explains the proposed approach with a step-by-step breakdown of the workflow. Section 4 introduces the performance metrics utilized for evaluating the effectiveness of abundance maps and endmember extraction using AEGEM. Section 5 presents the benchmark datasets used in the study. Section 6 demonstrates the superior performance of AEGEM with the results and discussions. The conclusions are summarized in Section 7, and Finally, the future work perspectives are presented in Section 8.

\section{Related work}

In this section, preliminary concepts relevant to the main theme are reviewed, highlighting the state-of-the-art. The analysis encompasses definitions of hyperspectral images, spectral unmixing, and presents important theorems and an exploration of the advantages and disadvantages of the current SU methods. Foundational concepts regarding graph convolutional neural networks are presented, as well.

\subsection{Hyperspectral Images}

HSI comprises data recorded in hundreds of narrow bands within the wavelength range of 500-2500 nm, which may vary depending on the sensor used. This high spectral resolution empowers material analysis in the image through the examination of spectral signatures \cite{Zheng2022}. Typically, HSI is represented as a 3D hypercube with dimensions denoted as $W\times H\times L$. where $W\times H$ corresponds to the width and height, representing the rows and columns of the image, respectively, while $L$ indicates the number of bands. 

\subsection{Spectral Unmixing}

Hyperspectral images consist of multiple contiguous narrow bands, making them suitable for performing spectral unmixing to analyze material composition. This involves extracting spectral signatures, known as endmembers, and fractional abundance maps representing the percentage of each pixel's composition in an acquired scene. Spectral unmixing is typically addressed within the framework of linear mixing models, as described in Equation \ref{LMM}.

\begin{equation}
X_{n} = \mathbf{M_{0}}\mathbf{\alpha}_{n} + \mathbf{\eta}_{n}\\
\mathbf{1}^{T}\mathbf{\alpha}_{n}\\
\alpha_{n}\geq 0
\label{LMM}
\end{equation}

Here, $X_{n}$ represents a given pixel with $L$ spectral bands, $M_{0}$ is an $L\times P$ matrix, where $P$ represents the endmembers, and $\alpha_{n}$ is the vector containing the abundances maps for each endmembers in pixel $X_{n}$, the vector $\eta_{n}$ represents additive noise. On the other hand, SU has been addressed by geometric methods, such as the volume maximization problem of a convex surface where the vertices are the endmembers or pure materials, and the mixed pixels are contained inside the convex hull. Approaches like vertex component analysis \cite{Nascimento2005} have been applied in this context. Currently, SU has been explored using deep learning approaches, primarily based on blind convolutional autoencoders \cite{Palsson2022}. The proposed models exhibit variations in the number of layers; typically, the encoder performs abundance map estimation, and the decoder handles endmember extraction. However, some architectures focus on only one of these problems; in \cite{Tulczyjew2022}, fractional abundance map estimation is performed. Additionally, in \cite{Ranasinghe2020}, a model is proposed that handles endmembers extraction in the encoder stage and the fractional abundances map estimation in the decoder stage.

\begin{figure*}[ht!]
    \centering
    \includegraphics[width=1.0\linewidth]{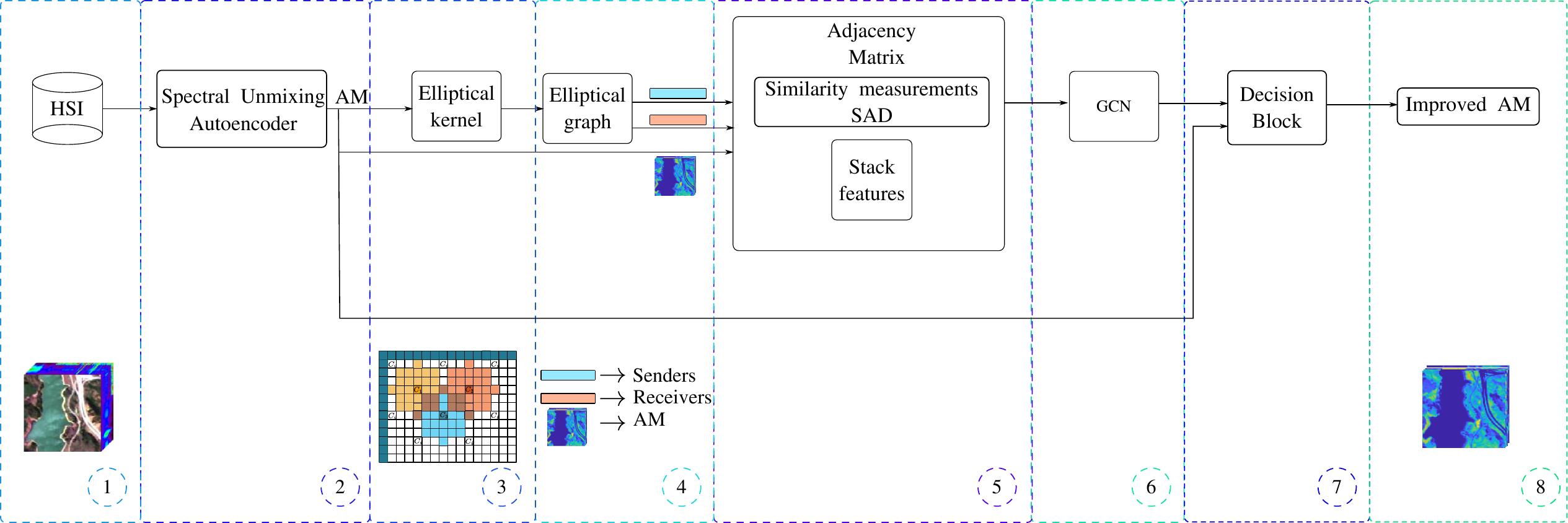}
    \caption{The spectral unmixing workflow designed for extracting endmembers and fractional abundance maps consists of eight stages. First, the HSI is loaded. Subsequently, a convolutional autoencoder is employed to extract endmembers and fractional abundances maps. An elliptical kernel is then applied to construct an elliptical graph, where centroids serve as senders and the remaining pixels within the geometry act as receivers to measure spectral distances, thereby generating the adjacency matrix within the elliptical neighborhood. The next step involves stacking abundance maps, senders, and receivers as inputs for the GCN. The GCN processes this input to refine abundance maps. Finally, an ensemble decision-making process determines the best abundance maps based on the RMSE metric.}
    \label{proposed_pipeline}
\end{figure*}

\subsection{Graph Definition}

A graph is a data structure used to represent complex objects in a non-Euclidean space, establishing relationships among a collection of objects and their respective properties. In this context, the objects are referred to as nodes, and the properties or features as edges. The formal definition specifies that a graph can be denoted as $G=\left(\nu, \varepsilon  \right)$, where $\nu$ is a set of nodes, and $\varepsilon$ is a set of edges connecting these nodes. In hyperspectral images, the nodes can be associated with pixels, and the edges are determined by similarity measurements \cite{Wan2021}-\cite{Liu2020} within a local neighborhood.

\subsection{Adjacency Matrix}

A graph, denoted as $G=\left(\nu, \varepsilon\right)$, can be represented by an adjacency matrix, where $\mathbf{A}_{i,j}$ establishes connections between nodes $\nu_{i}$ and $\nu_{j}$. If nodes $\nu_{i}$, and $\nu_{j}$ are adjacent, then the position $\mathbf{A}_{i,j}$ is set to the value obtained from Equation \ref{RBF}; Otherwise, it is set to 0. Typically, the adjacency matrix is obtained using the radial basis function to determine similarity in a neighborhood, as depicted in Equation \ref{RBF}, where the numerator term $-\left\|\mathbf{x}_{i}-\mathbf{x}_{j}\right\|$ corresponds to the Euclidean distance between $\mathbf{x}_{i}$ and $\mathbf{x}_{j}$, $\sigma$ represents the width of the radial basis function. 
\begin{equation}
    \label{RBF}
	\mathbf{A}_{\mathbf{(i,j)}}  = \exp{\left(\frac{-\left\|\mathbf{x}_{i}-\mathbf{x}_{j}\right\|}{\sigma^{2}}\right)}
\end{equation}

Another matrix representation for graphs is the Laplacian matrix. Given the adjacency matrix $\mathbf{A}$ and the degree matrix $\mathbf{D}$, the Laplacian is given by Equation \ref{laplacian_matrix}, and the normalized Laplacian is given in Equation \ref{norm_laplacian}, where $\mathbf{I}$ is the identity matrix.

\begin{equation}
    \label{laplacian_matrix}
    \mathbf{L} = \mathbf{D}-\mathbf{A}
\end{equation}

\begin{equation}
\label{norm_laplacian}
	\mathbf{L}= \mathbf{D}^{-\frac{1}{2}}\left(\mathbf{D}-\mathbf{A}\right)\mathbf{D}^{-\frac{1}{2}}\\
  = \mathbf{I} - \mathbf{D}^{-\frac{1}{2}}\mathbf{A} \mathbf{D}^{-\frac{1}{2}}
\end{equation}

\subsection{Graph Convolutional Neural Networks}

Given the input data $Y$, the spectral convolution can be defined as follows:
\begin{equation}
    \label{gcn_1}
    \theta \ast Y = \theta\left(LY\right)\\
    =\theta \left(U \Lambda U^{T}\right)Y\\
    = U \Theta U ^{T} Y
\end{equation}
Here, $\ast$ represents the graph convolutional operation, $\theta$ is the convolutional kernel, $\Theta$ is the filter in the spectral domain, and $U$ is the spectral Fourier basis containing the eigenvectors of the Laplacian matrix. $\Lambda$ is the diagonal matrix containing the eigenvalues derived from the decomposition of $\mathbf{L}$ \cite{Gao2022}. The decomposition matrix is described in the Equation \ref{norm_laplacian}, \ref{gcn_1}, where $\mathbf{I}$ is the identity matrix, $\mathbf{D}$ is the diagonal degree matrix, and $\mathbf{A}$ is the adjacency matrix. The kernel $\theta$ can be rewritten as follows:

\begin{equation}
    \Theta\left(\Lambda ^{N}\right) = \sum_{n=0}^{N-1}\theta_{n}\Lambda^{n}
\end{equation}

Then, formulating the kernel $\theta$ as a polynomial of order $N$, it can be expressed as:

\begin{equation}
    \theta \ast Y = \phi\left(\Theta\left(\Lambda^{n}\right)Y\right)
\end{equation}

To perform the approximation of the first Chebyshev polynomial, the graph convolutional operation is defined as:

\begin{equation}
    H\left(Y, W\right) = \phi\left(D^{-\frac{1}{2}}\hat{A}D^{-\frac{1}{2}}Y W\right)
\end{equation}

Here, $\hat{A}$ is the adjacency matrix with self-connection, $W$ corresponds to trainable weights, and $\phi$ is the ReLU activation function.

\begin{figure*}[ht!]
    \centering
    \includegraphics[width = 0.9\linewidth]{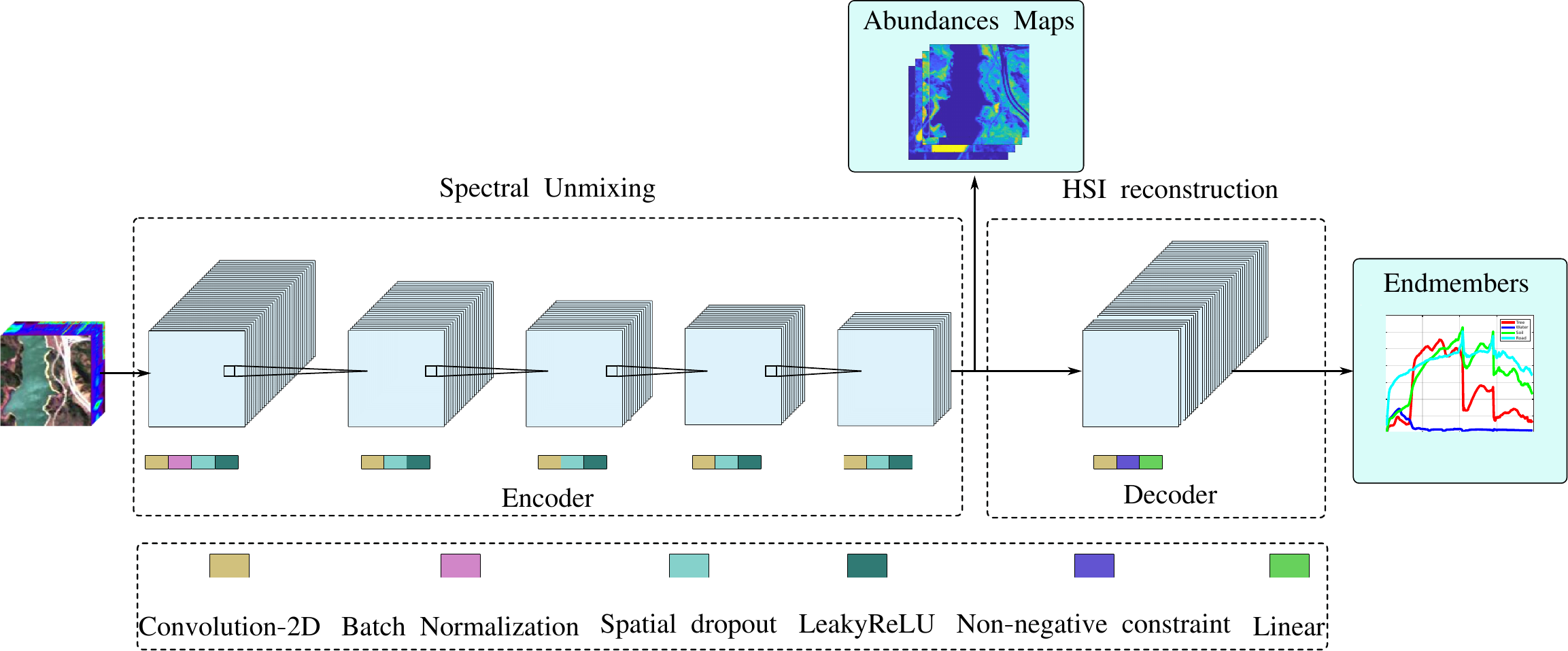}
    \caption{Architectural Illustration of the convolutional autoencoder with its respective operations and activation functions used to perform spectral unmixing analysis. In the encoder, abundance maps estimation is conducted, while in the decoder, endmembers extraction is performed.}
    \label{autoencoder_su}
\end{figure*}

\section{Proposed approach}

In order to address the problem of SU of hyperspectral images, the AEGEM workflow is depicted in Figure. \ref{proposed_pipeline} is proposed comprising of eight stages. The process commences with the HSI as the input data. The HSI is denoted as follows:

\begin{equation}
\mathbf{Z}= 
    \begin{bmatrix}
    Z_{1}^{1}&Z_{1}^{2}&Z_{1}^{3}&\ldots&Z_{1}^{B}\\
    Z_{2}^{1}&Z_{2}^{2}&Z_{2}^{3}&\ldots&Z_{2}^{B}\\
    Z_{3}^{1}&Z_{3}^{2}&Z_{3}^{3}&\ldots&Z_{3}^{B}\\
    \vdots&\vdots&\vdots&\ddots&\vdots\\
    Z_{L}^{1}&Z_{L}^{2}&Z_{L}^{3}&\ldots&Z_{L}^{B}\\
    \end{bmatrix}
\end{equation}

The HSI has dimensions $H\times W \times L$, representing height, width and the number of bands, where $B =H \times W$. SU analysis is performed as a preprocessing stage using a convolutional autoencoder, designed to extract endmembers and abundances maps, as depicted in Figure. \ref{autoencoder_su}. The autoencoder is an unsupervised deep learning architecture composed of an encoder and a decoder. The encoder transforms the input data into a hidden representation, denoted as $\mathbf{f_{i}} = F_{E}\left(\mathbf{x_{i}}\right)$. Subsequently, the decoder, given by $\hat{\mathbf{x_{i}}} = F_{D}\left(\mathbf{f_{i}}\right)$, reconstructs the data, subject to a loss function $L\left(\mathbf{x_{i}}, F_{D}\left(F_{E}\left(\mathbf{x_{i}}\right)\right)\right)$, in order to conduct accurate reconstruction, it is necessary to add a regularization function given by $L=L\left(\mathbf{x_{i}}, \mathbf{\hat{x_{i}}}\right)+ \gamma \varsigma \left(\mathbf{f_{i}}, \mathbf{W_{e}}, \mathbf{W_{d}}\right)$ where $\gamma$ is a tuning parameter, and $\varsigma\left(\mathbf{f_{i}}, \mathbf{W_{e}}, \mathbf{W_{d}}\right)$ is the penalty function. 

The reconstructed data obtained from the decoder can be expressed as a result of activation functions $\phi_{e}, \phi_{d}$ and weighted matrices $\mathbf{W_{e}}$, $\mathbf{W_{d}}$ from both the encoder and decoder, as shown in Equation \ref{auto_act}.

\begin{equation}
    \label{auto_act}
    \hat{\mathbf{x_{i}}} = \phi_{D}\left(\mathbf{W_{d}}\left(\phi_{E}\left(\mathbf{W_{e}}\right)\right)\right)
\end{equation}

Then the reconstructed HSI can be expressed as shown in Equation \ref{auto_decom}, where $\phi^{N-1}$ represents the activations from the $\left(N-1\right)$ layer, and $\left(N-1\right)^{s}$ denotes the $s$ number of neurons in that layer. In a compact way, the reconstruction performed in the decoder can be written as shown in Equation. \ref{decoder_compact}

\begin{equation}
\label{auto_decom}
\mathbf{\hat{X}_{i}} = 
    \mathbf{W}_{d}^{N}
    \begin{bmatrix}
        \phi^{(N-1)(1)}&\phi^{(N-1)(2)}&\ldots&\phi^{(N-1)(w)}  
    \end{bmatrix}
\end{equation}

\begin{equation}
\label{decoder_compact}
    \mathbf{\hat{X_{i}}} = \mathbf{W}_{d_{B}}^{N} \mathbf{\phi}_{P_{s}}
\end{equation}

In the third stage, an elliptical kernel is described by the canonical Equation. \ref{elliptical_kernel} is applied to the abundance maps from the previous step. The semi-major and semi-minor axes are set to form a neighborhood mask of $a\times b \times L$, where $a$, $b$ are the number of pixels in the semi-minor and semi-major axes, respectively, and $L$ is the number of bands, determined based on the results obtained in the abundances maps for Samson, Jasper, and Urban datasets. 

\begin{equation}
\label{elliptical_kernel}
    \frac{\mathbf{x^{2}}}{a^{2}} + \frac{\mathbf{y^{2}}}{b^{2}} + \frac{\mathbf{z^{2}}}{L^{2}} = 1
\end{equation}

\begin{figure}
    \centering
    \includegraphics[width = 0.75\linewidth]{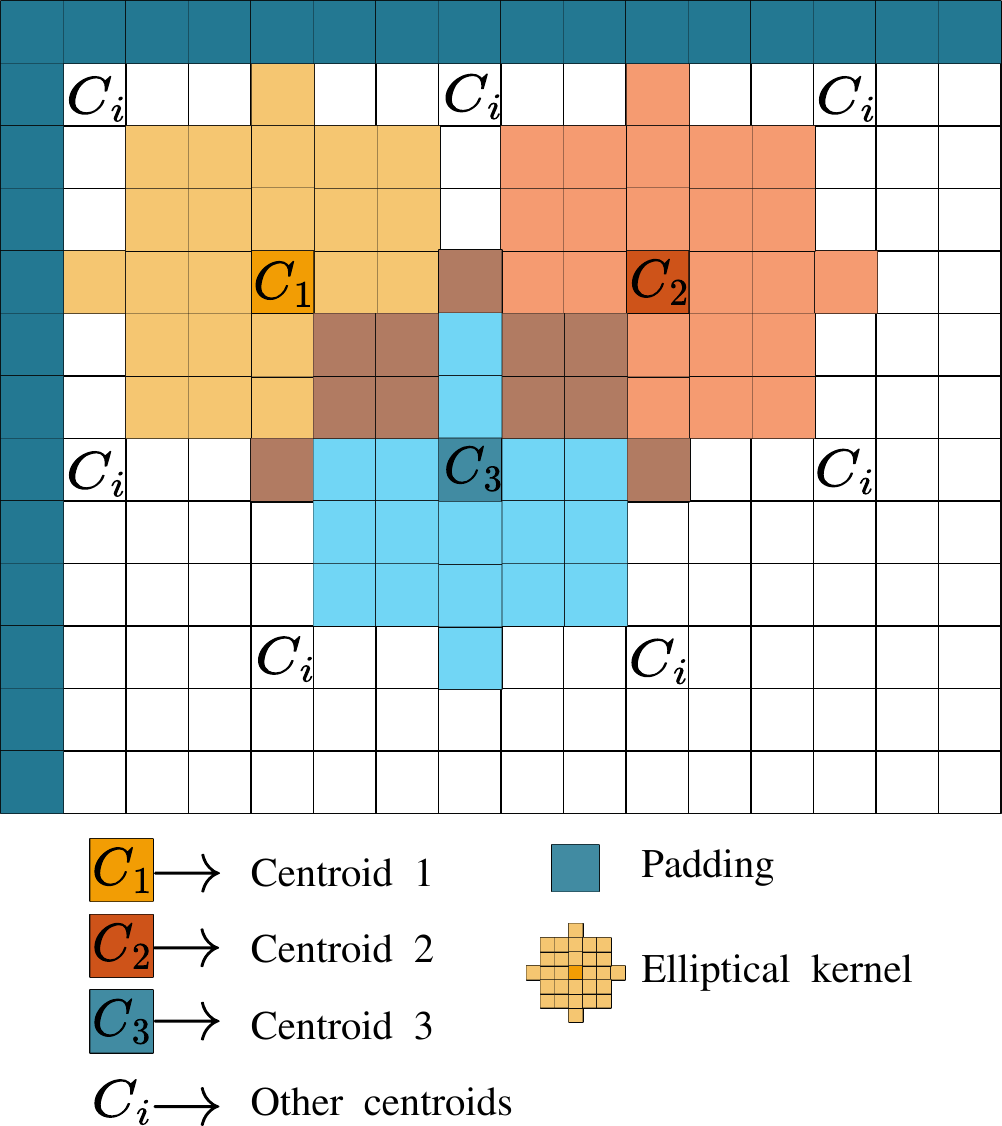}
    \caption{Illustration of the elliptical kernel with the locations of their respective centroids. The intersections in each ellipse are highlighted in brown. This elliptical kernel is used to conduct similarity measurements for the elliptical graph in the GCN.}
    \label{kernel_representation}
\end{figure}

After the elliptical kernel application, an elliptical graph is constructed based on a star topology, as illustrated in Figure. \ref{kernel_representation}, where the central pixel serves as the sender and the remaining pixels in the neighborhood act as receivers. Subsequently, the adjacency matrix is formed using the spectral angle distance, as described in Equation \ref{adj_sad}, where the $\mathbf{x_{i}}$ represents the central pixel in the elliptical kernel, and $\mathbf{x_{j}}$ denotes the surrounding pixels in the mask. The parameter $N$ corresponds to the number of samples from the remaining pixels within the elliptical kernel, including the centroid. Furthermore, features from the abundance maps for both senders and receivers are stacked. For each pixel in the image, we obtain representation of $p_{s}$ senders and $p_{r}$ receivers, extracted from the abundance maps based on the number of endmembers in the image. Specifically, $p_{s} = [f_{1},f_{2},f_{3},\ldots,f_{P}]$, and $p_{r} = [f_{1},f_{2},f_{3},\ldots,f_{P}]$

\begin{equation}
\label{adj_sad}
    A_{i,j} = \frac{1}{N} \sum_{i=1}^{N}\arccos\left(\frac{<{\mathbf{x_{i}}}, \mathbf{x_{i}}>}{\left \| {\mathbf{x_{i}}}\right \|_{2}\left \| {\mathbf{x_{j}}}\right \|_{2}}\right) 
\end{equation}

\begin{figure*}
    \centering
    \includegraphics[width = 0.85\linewidth]{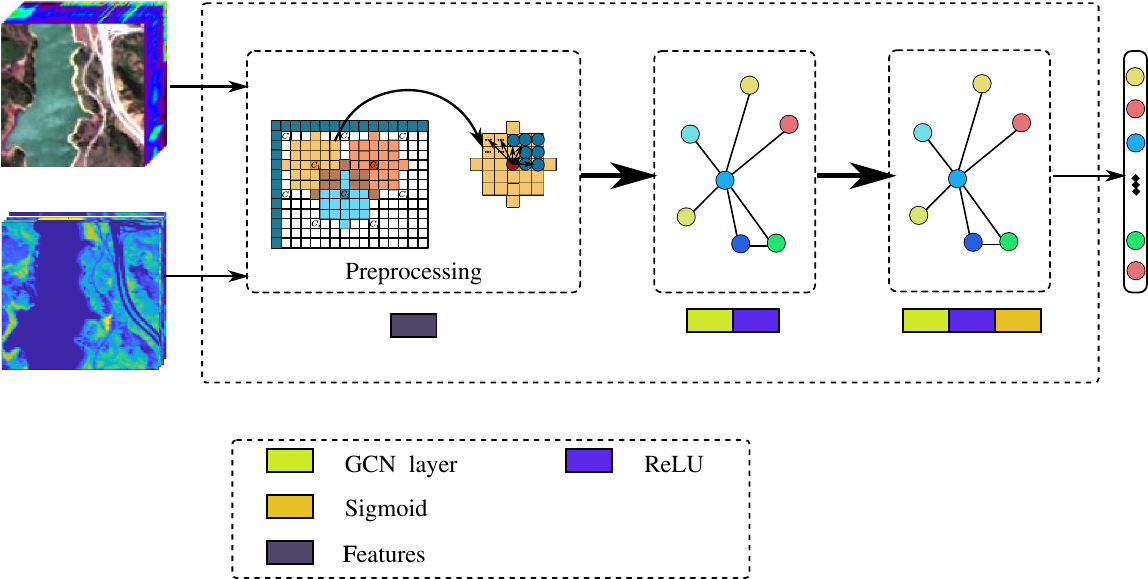}
    \caption{Architectural illustration of the GCN, showcasing the operations conducted and the activation functions used for abundance map extraction. The depiction includes the preprocessing stage with the HSI, the abundance maps, and the ground truth.}
    \label{GCN_configuration}
\end{figure*}

The adjacency matrix, along with the stacked features extracted from the abundance maps originally generated by the autoencoder in the second stage of the proposed workflow (see Figure. \ref{proposed_pipeline}), and the labels from the ground truth are utilized to train the GCN for reconstructing the abundance maps. Subsequently, a decision-making block is employed to determine the best abundance maps based on the RMSE metric.

The pseudocode for the autoencoder, data processing (referred to as elliptical operations), and GCN are described in Algorithms \ref{alg:autoencoder}, \ref{alg:elliptical}, and \ref{alg:gcn}, respectively. The pseudocode in Algorithm \ref{alg:autoencoder} details the input and output of the autoencoder. Pseudocode \ref{alg:elliptical} presents the steps for the proposed elliptical kernel, which enhances the features from the HSI and the abundance maps obtained from the autoencoder. Finally, pseudocode \ref{alg:gcn} describes the application of the GCN, which processes the enriched features to improve the final abundance maps. By leveraging the strengths of each component, the autoencoder effectively extracts initial abundance maps and endmembers, the elliptical kernel enhances these maps, and the GCN refines the results, leading to superior performance of AEGEM on the benchmark datasets. 

The AEGEM is executed on a Dell Precision Server 7920 Rack with an Intel Xeon Gold processor, a 4 GB NVIDIA T1000 graphics card, a 1 TB SATA hard drive, and 64 GB RAM. In terms of computational cost, the proposed AEGEM primarily incurs computational cost from matrix multiplications due to its convolutional operations derived from the convolutional autoencoder. This complexity arises during both the preprocessing step and the application of the elliptical kernel and the GCN. According to big O notation, this yields $O (n m p + n^{2})$ where $n$ is the number of samples, $m$ is the input dimension, and $p$ is the output feature dimension

\begin{algorithm}[h]
\caption{\textbf{Autoencoder}}\label{alg:autoencoder}
\begin{algorithmic}
\State \textbf{Input:} $\mathbf{Z}$ 
\State $\mathbf{Z}$ $\gets$ HSI 
\State $\mathbf{P}$ $\gets$ Endmembers
\State $\alpha \gets $Abundance maps
\State $\mathbf{P}, \mathbf{\alpha}$ $\gets$ \textbf{Equation}. \ref{auto_act}
\State \textbf{Output:} $\mathbf{P}, \mathbf{\alpha}$
\end{algorithmic}
\end{algorithm}

\begin{algorithm}[h]
\caption{\textbf{Elliptical operations}}\label{alg:elliptical}
\begin{algorithmic}

\State \textbf{Input:}
\State $major \gets $Semi-major axis
\State $minor \gets $Semi-minor axis
\State $\mathbf{P}$ $\gets$ Endmembers
\State $\alpha \gets $Abundance maps
\State $\mathbf{Z}$ $\gets$ HSI 
\State \textbf{Elliptical kernel:}
\State $mask \gets $Elliptical mask \textbf{Equation }\ref{elliptical_kernel}
\State $c \gets $Centroids in HSI \textbf{Equation }\ref{elliptical_kernel}
\State $n(c) = \mathbf{Z} \times mask(c)$ $\gets$ \Comment{neighborhood for each centroid in the image}
\State \textbf{Elliptical graph:}
\State $S(c, pixel) = SAD(c, n(c, pixel))$ $\gets$ \Comment{Compute SAD for each pixel in the neighborhood}
\State \textbf{Adjacency matrix:}
\State $M(c, pixel) = stack(P(c), \alpha(c), P(pixel), \alpha(pixel), S(c, pixel))$ $\gets$ Stack of features
\State \textbf{Output:} M
\end{algorithmic}
\end{algorithm}

\begin{algorithm}[h]
\caption{\textbf{GCN}}\label{alg:gcn}
\begin{algorithmic}
\State \textbf{Input:} \textbf{M} 
\State Train GCN
\State GCN production
\State \textbf{Output:} Improved \textbf{P}
\end{algorithmic}
\end{algorithm}

\subsection{Hyperparameters configurations for the convolutional autoencoder}
The AEGEM consists of two deep learning models: a convolutional autoencoder and GCN. The convolutional autoencoder is implemented in Python using TensorFlow libraries. It comprises an encoder and a decoder. The encoder is responsible for extracting the abundance maps and consists of five 2-dimensional convolutional layers, as shown in Table \ref{autoencoder_details}. For each layer is applied the following filter sizes, respectively, 5, 3, and P, where P corresponds to the number of endmembers present in the acquired scene. The choice of smaller filter sizes is better because increasing the filter size negatively impacts the abundances maps, introducing noise at the borders and causing a blurring effect. The abundance maps are then normalized using the sum-to-one (ASC) constraint performed by a softmax function with a scaling factor of 5.

In the decoder stage, endmembers extraction and HSI image reconstruction are performed. Non-negative constraints are applied, along with linear operations. The unique 2-dimensional convolutional operation in the decoder has L filters, corresponding to the number of bands, and a filter size of 7 is used.

\subsection{Hyperparameters configuration for the GCN}

The GCN is programmed in Python using the PyTorch libraries. In this case, GCN is utilized to enhance the extracted abundance maps generated by the convolutional autoencoder. The parameters are fine-tuned using 10-fold cross-validation. The learning rate is set at $0.001$, with a total configuration of 200 epochs and 128 hidden nodes. The chosen optimizer is Adam, and the loss function employed is cross-entropy.

The GCN comprises two convolutional layers, as depicted in Figure. \ref{GCN_configuration}; initially, a convolutional layer with a number of connections corresponding to the edges is applied, followed by a ReLu activation function. Subsequently, a second GCN layer with a specified number of filters is employed, followed by a sigmoid function to enforce and ensure adherence to the ASC constraints, thereby enhancing the extraction of abundance maps.
\begin{table}[!t]
\renewcommand{\arraystretch}{1.3}
\caption{Configurations and parameters settings for the proposed Autoencoder.}
\label{autoencoder_details}
\centering
\begin{tabular}{|c||c|}\hline
\textbf{Parameters} & \textbf{Description}\\ \hline
\multicolumn{2}{|c|}{\textbf{Encoder}}\\ \hline
Input data & $9 \times 9$\\ 
2D convolution & 128\\
Filter size &5\\
2D convolution &64\\
Filter size & 3\\
2D convolution & 32\\
Filter size &3\\
2D convolution & P\\
Filter size &1\\
Sum to one constraint Softmax& 5\\ \hline
\multicolumn{2}{|c|}{\textbf{Decoder}}\\ \hline
Batch normalization& \\
2D convolution& L\\
filter size & 7\\ \hline
\end{tabular}
\end{table}
\section{Performance metrics}
\label{metrics}
In this section, the metrics used to assess the proposed model AEGEM for endmember extraction and the computation of fractional abundance maps are presented. $\mathbf{\hat{p}} = [\hat{p_{1}},\hat{p_{2}},\hat{p_{3}},\ldots,\hat{p_{L}}]^{T}$ represent the L-dimensional endmember vector compared to the ground truth endmember $\mathbf{p} = [p_{1},p_{2},p_{3},\ldots,p_{L}]^{T}$ using spectral angle distance as shown in Equation \ref{sad}. For the abundance maps, the Root Mean Square Error metric Equation. \ref{rmse} is utilized, where $\alpha_{i}$ is the abundance map of the $i$-th endmember, and $\hat{\alpha}_{i}$ corresponds to the $i$-th abundance map for the ground truth endmember. 

\begin{itemize}
    \item Spectral Angle Distance (SAD):
    \begin{equation}
        \label{sad}
        \text{SAD}_{i} = \arccos{\left(\frac{\mathbf{\hat{p_{i}}^{T}}\mathbf{p_{i}}}{\left \| \mathbf{\hat{p_{i}}}\right \|_{2}\left \| \mathbf{p_{i}}\right \|_{2}}\right)}
    \end{equation}
    \item Root Mean Square Error (RMSE):
    \begin{equation}
    \label{rmse}
        \text{RMSE} = \sqrt{\frac{1}{R}\sum_{i=1}^P\left \| \alpha_{i}-\hat{\alpha_{i}} \right \|}
    \end{equation}
\end{itemize}
\section{Experiments}
\subsection{Benchmark datasets}
\label{benchmark}
The experiments involves applying the proposed AEGEM to three different datasets, regarded as benchmark datasets for hyperspectral unmixing: Samson, Jasper Ridge, and Urban. A detailed description of each dataset used in the experiments is provided below:

\begin{figure*}[h!]
    \centering
    \includegraphics[width=0.8\linewidth]{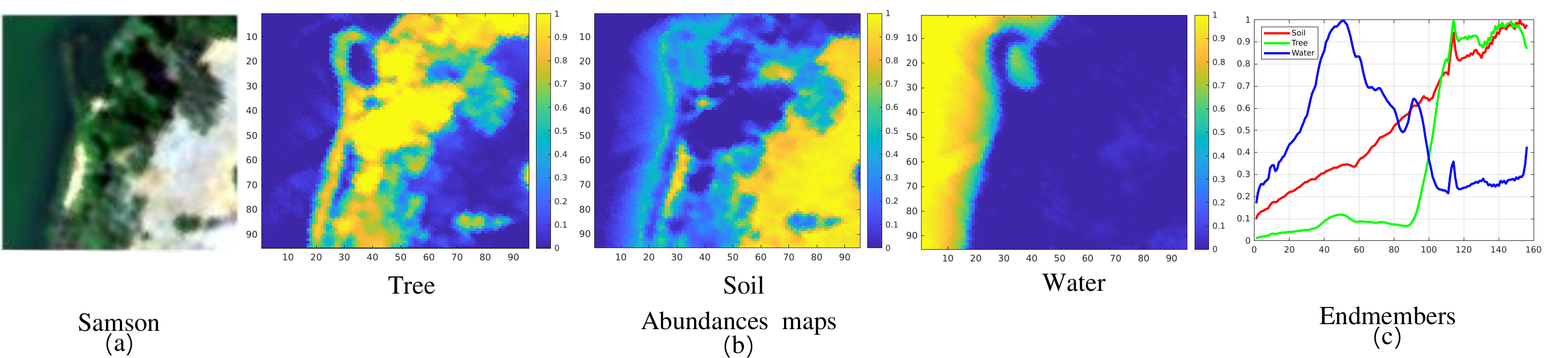}
    \caption{Samson dataset with corresponding endmembers and abundances maps for tree, soil, and water.}
    \label{samson_gt}
\end{figure*}

\begin{figure*}[h!]
    \centering
    \includegraphics[width=0.85\linewidth]{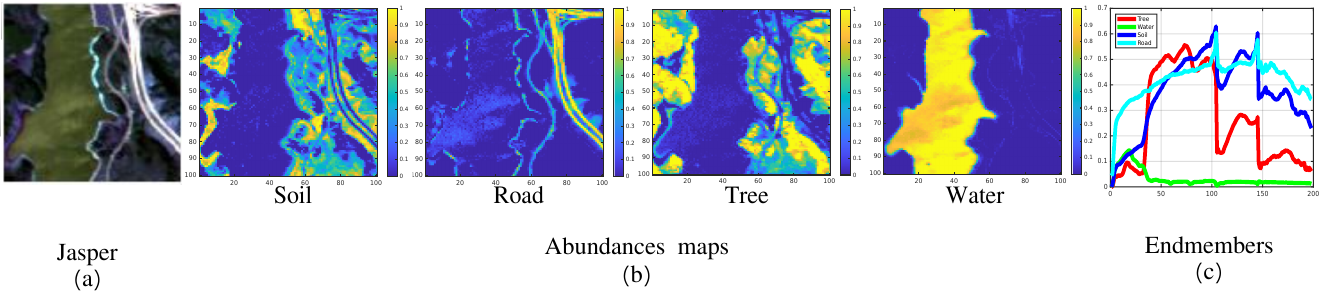}
    \caption{Jasper dataset with corresponding endmembers and abundances maps for soil, road, tree, and water.}
    \label{jasper_gt}
\end{figure*}

\begin{figure*}[h!]
    \centering
    \includegraphics[width=0.9\linewidth]{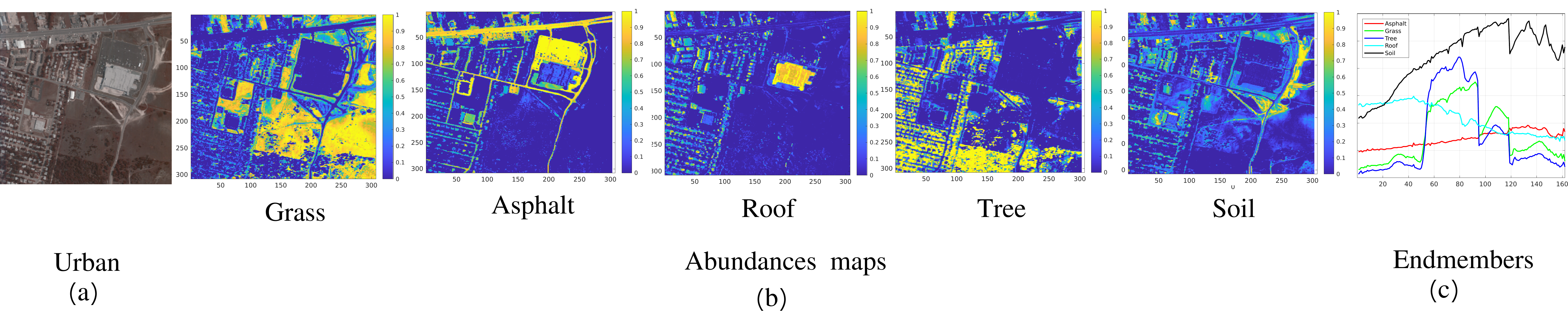}
    \caption{Urban dataset with corresponding endmembers and abundances maps for grass, asphalt, roof, tree, and soil.}
    \label{urban_gt}
\end{figure*}
\begin{itemize}
    \item Samson: Comprising $156$ bands, the Samson dataset records wavelengths between $401-889$ nm. The original image is cropped into three regions of interest, with this study focusing on ROI Samson $\#1$ \cite{Zhu2017HyperspectralUG}, having a spatial resolution of $95 \times 95$. Samson dataset is characterized by three endmembers: Soil, Tree, and Water, as illustrated in Figure. \ref{samson_gt}.
    
    \item Jasper: This dataset comprises $224$ bands with a spectral resolution of $9.46$ nm, recorded within the wavelength range of $380-2500$ nm. After applying noisy channel correction to account for atmospheric effects, 194 bands are selected. The region of interest spans $100 \times 100$ pixels, and the dataset encompasses four endmembers: Road, Soil, Water, and Tree, as depicted in Figure. \ref{jasper_gt}.

    \item Urban: This dataset consists of $210$ bands, and after the removal of channels affected by dense water vapor and atmospheric effects, $162$ bands remain. Recorded within a wavelength range of $400-2500$ nm, the spectral resolution is $307 \times 307$. For this study, datasets with with five endmembers are analyzed: Grass, Asphalt, Roof, Tree, and Soil, as depicted in Figure .\ref{urban_gt}.
\end{itemize}

\section{Results and discussions}
This section presents and discusses the results of the \textbf{AEGEM}: Autoencoder Graph Ensemble Model.
\subsection{Abundances maps estimation and endmember extraction from benchmark datasets: Samson, Jasper Ridge, and Urban}

The performance of AEGEM has been validated on three benchmark datasets: Samson, Jasper Ridge, and Urban, as described in Sub-section \ref{benchmark}. The metrics applied for assessing the endmembers and abundance maps are spectral angle distance (SAD) and root mean square error (RMSE), respectively, as detailed in Section \ref{metrics}. AEGEM's performance is ranked in comparison with the performance of the state-of-the-art models CNNAEU \cite{Palsson2021}, UnDIP \cite{Rasti2022}, and SGSNMS \cite{Wang2017}. The results show that AEGEM improves the estimation of abundance maps for large extended regions due to the ensemble solution, which refines the abundance maps obtained from the autoencoder. The processing steps, including the application of the elliptical kernel and the subsequent graph construction from the spectral angle distance feature provide an enriched input to the GCN. Specifically, for the Samson and Jasper datasets, water has the best reconstruction, while for the Urban dataset, the roof and asphalt show the best reconstruction.

\begin{figure*}[h!]
    \centering
    \includegraphics[width=0.7\linewidth]{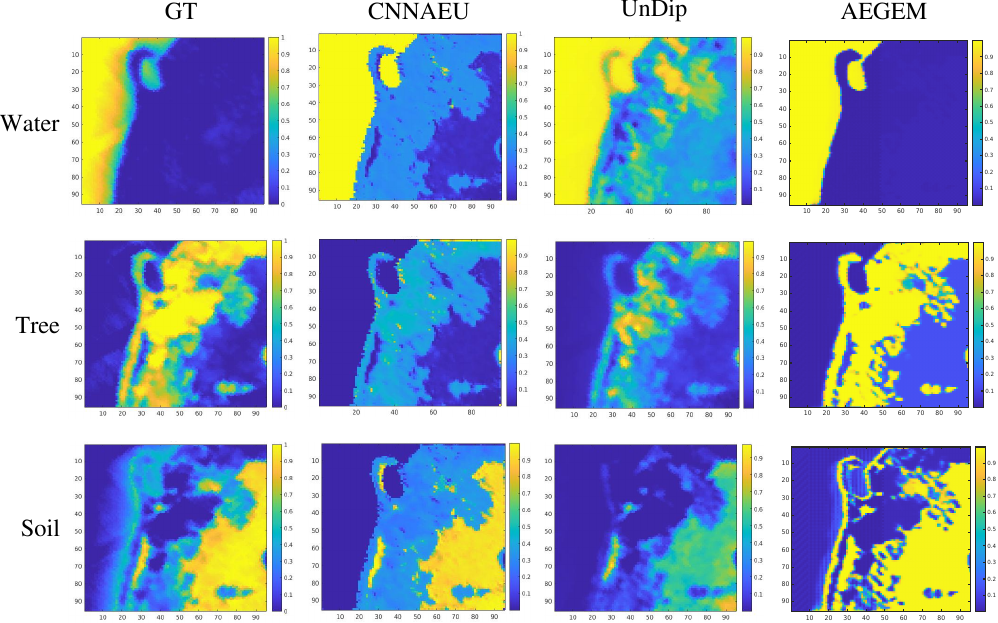}
    \caption{Comparison of abundances maps among CNNAEU, UnDIP, and the proposed model AEGEM for the Samson dataset, illustrating the presence of three materials: Water, Tree, and Soil.}
    \label{samson_maps}
\end{figure*}
\begin{figure*}[h!]
    \centering
    \includegraphics[width=0.65\linewidth]{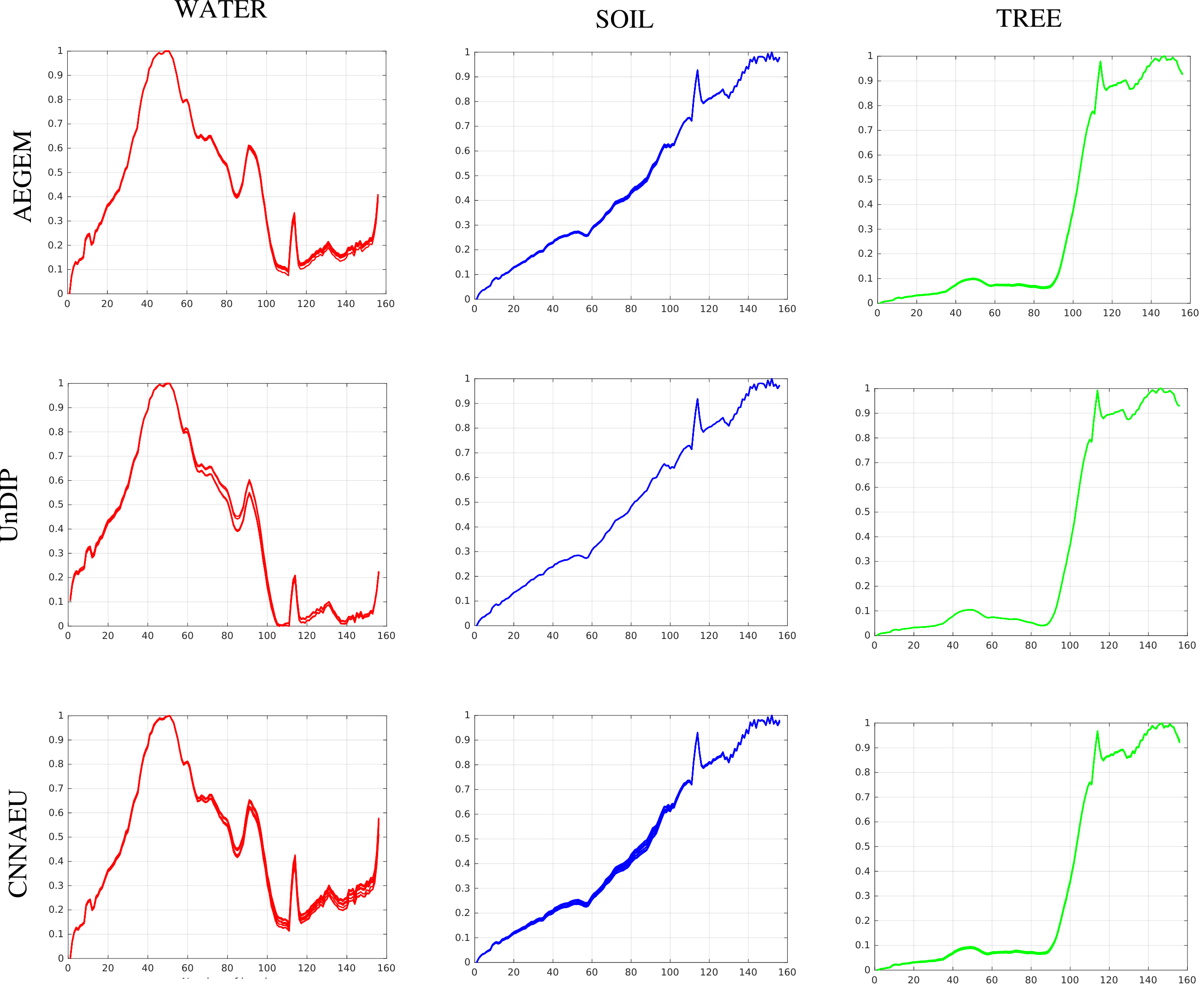}
    \caption{A comparative analysis of endmember extraction over the Samson dataset illustrates water, soil, and tree extraction performed by CNNAEU, UnDIP, and the proposed model AEGEM, conducted over 10 runs.}
\end{figure*}

\subsection{Results with Samson dataset}

\begin{figure*}[!h]
\centering
    \includegraphics[width=0.7\linewidth]{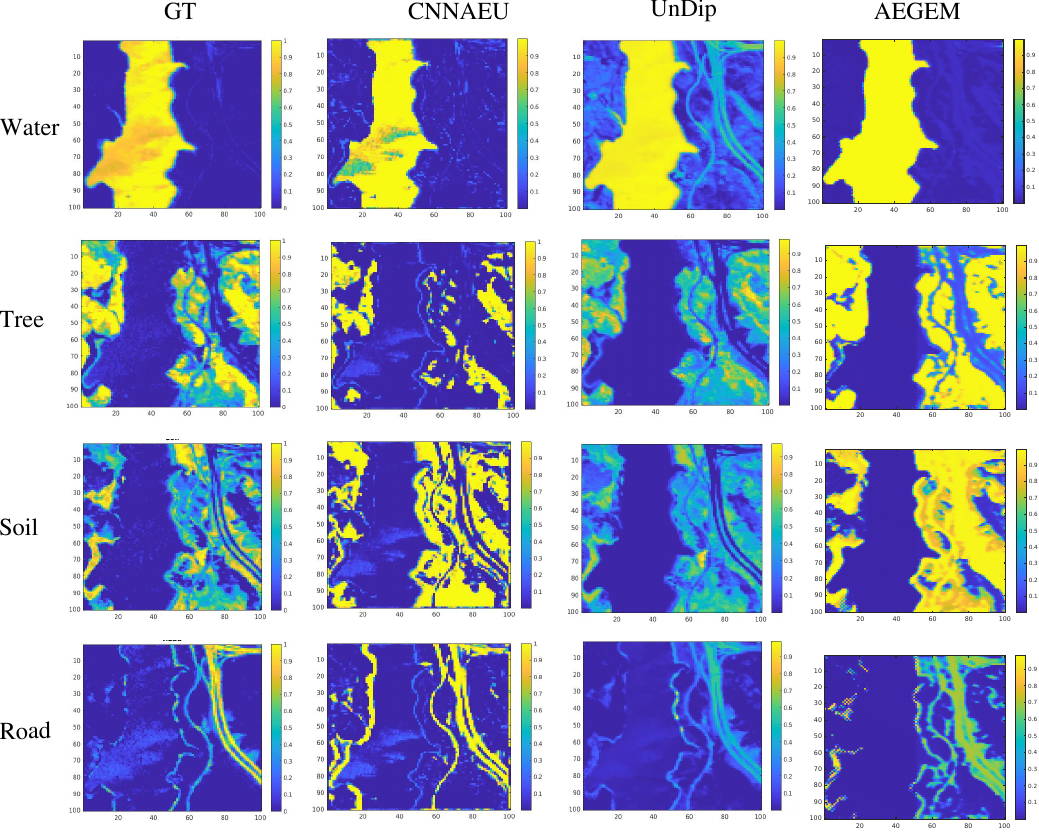}
    \caption{Comparison of abundances maps among CNNAEU, UnDIP, and the model AEGEM for the Jasper Ridge dataset, illustrating the presence of three materials: Water, Tree, Soil, and Road.}
    \label{jasper_maps}
\end{figure*}

The Samson dataset described in Section \ref{benchmark}, comprises three endmember: Tree, Soil and Water. To evaluate the proposed model AEGEM, experiments are conducted ten times to facilitate statistical analysis. AEGEM consistently outperforms baseline algorithms in abundance map extraction, yielding SAD values of $0.158$ for Tree, $0.182$ for Soil, and $0.081$ for Water. Additionally, on endmember estimation, the Water endmember exhibited superior results compared to baseline models, demonstrating the lowest average for both RMSE and SAD. The comparison of the abundance maps is depicted in Figure. \ref{jasper_maps}, and the results for the endmembers and abundances maps for the baseline algorithms and the proposed model AEGEM are given in Table. \ref{samson_description}.

\subsection{Results with Jasper Ridge dataset}

The Jasper Ridge dataset comprises four endmembers- Tree, Soil, Water, and Road respectively.  The AEGEM, when compared with baseline algorithms, outperforms the unmixing results obtained for Water with an SAD of $0.110$. The endmembers extraction also exhibits superior performance with SAD values of $0.035$ for Tree, $0.060$ for Water, and the lowest average SAD of $0.109$. For the Urban dataset, AEGEM achieves the best performance in estimating abundance maps for Roof and Asphalt, obtaining $0.135$, $0.240$ respectively utilizing the RMSE metric compared to the ground truth, the comparison of the abundances maps is depicted in Figure. \ref{urban_comp}. Conversely, for the endmembers corresponding to grass and roof, AEGEM extracted endmembers outperforms using the SAD metric, obtaining respective values of $0.063$ and $0.094$. The comparative analysis over ten runs is depicted in Figure. \ref{urban_endmember}.

\subsection{Results with Urban dataset}

\begin{figure*}[h!]
    \label{urban_comp}
    \centering
    \includegraphics[width = 0.75\linewidth]{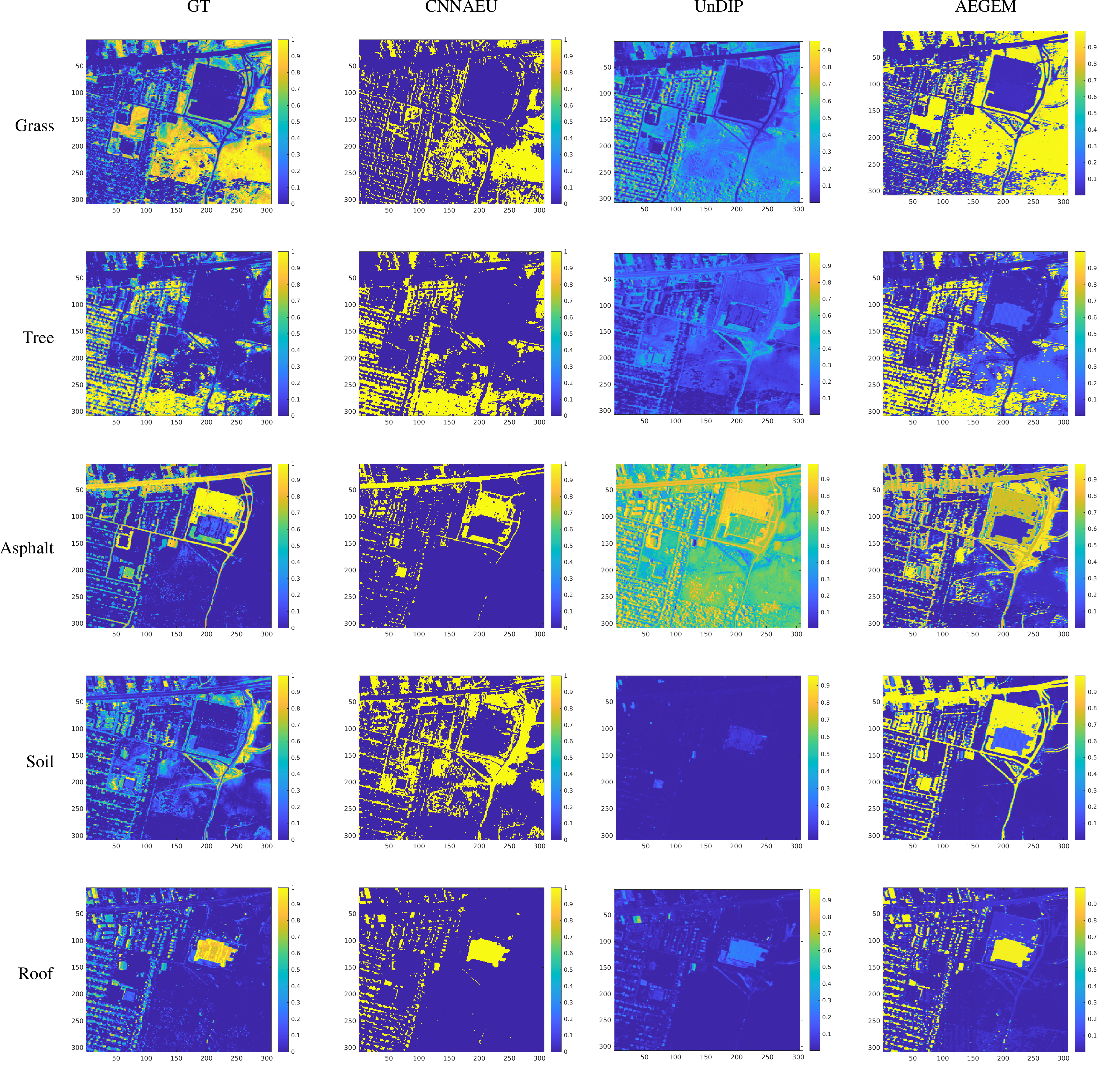}
    \caption{Comparison of abundances maps among CNNAEU, UnDIP, and the proposed model AEGEM for the Urban dataset, illustrating the presence of four materials: Grass, Tree, Asphalt, Soil, and Water.}
\end{figure*}

\begin{figure*}[ht!]
    \label{urban_endmember}
    \centering
    \includegraphics[width = 0.85\linewidth]{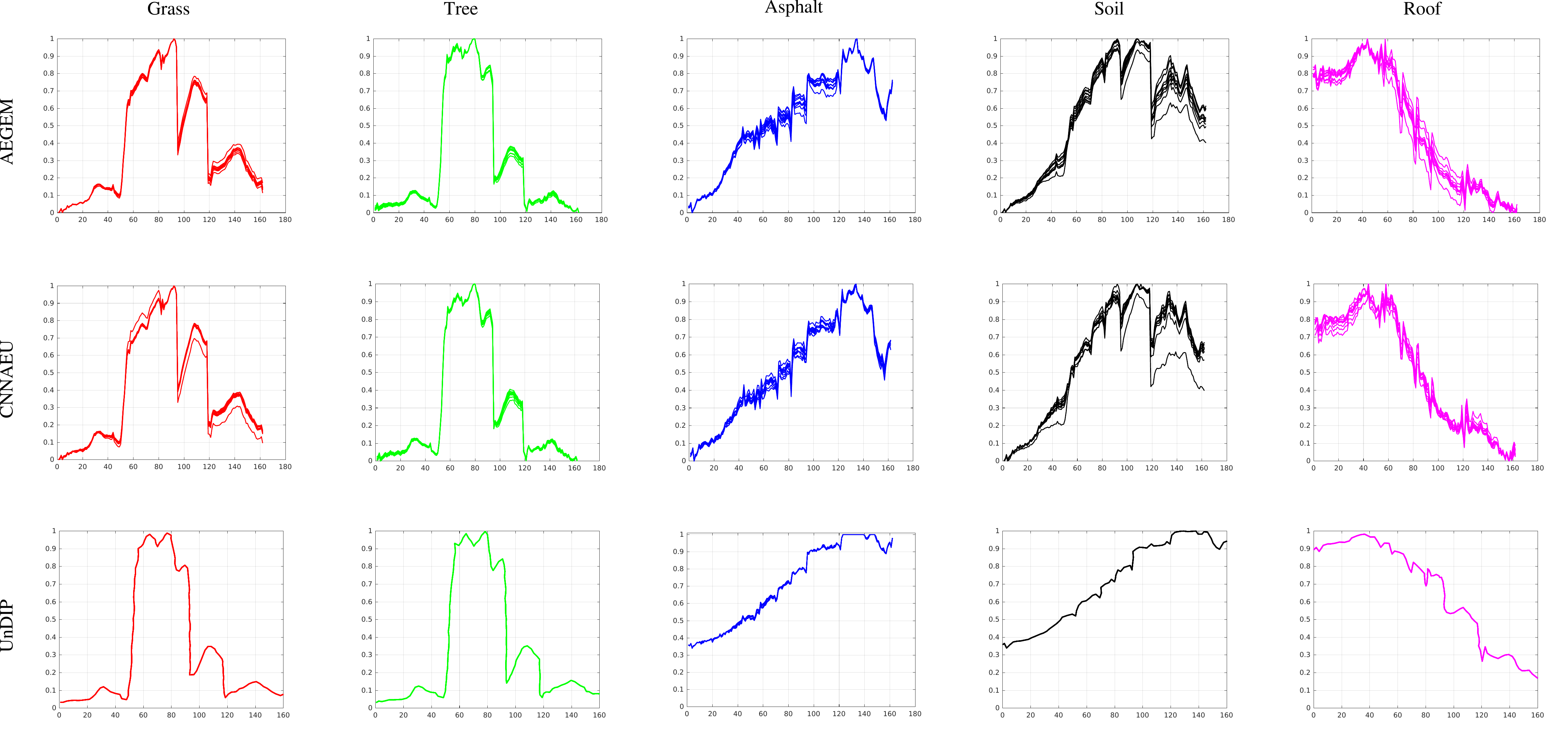}
    \caption{A comparison of spectral endmembers extracted from the Urban dataset illustrating grass, tree, asphalt, soil, and roof endmembers.}
\end{figure*}

Proceeding with the analysis for the Urban dataset, the experiments are run ten times, and the average measurements for each material in the image are reported in Table .\ref{urban_description}. Additionally, Figure. \ref{urban_comp} compares the reconstructed abundance map achieved by the chosen baseline algorithms and AEGEM, while the endmembers are depicted in Figure. \ref{urban_endmember}. The AEGEM achieves better results in the abundance maps, for roof and asphalt, with values of $0.135$ and $0.240$, respectively. For the endmembers, AEGEM achieved better results for grass and roof, with values of $0.063$ and $0.094$, respectively.

\begin{figure*}[!h]
    \centering
    \includegraphics[width = 0.7\linewidth]{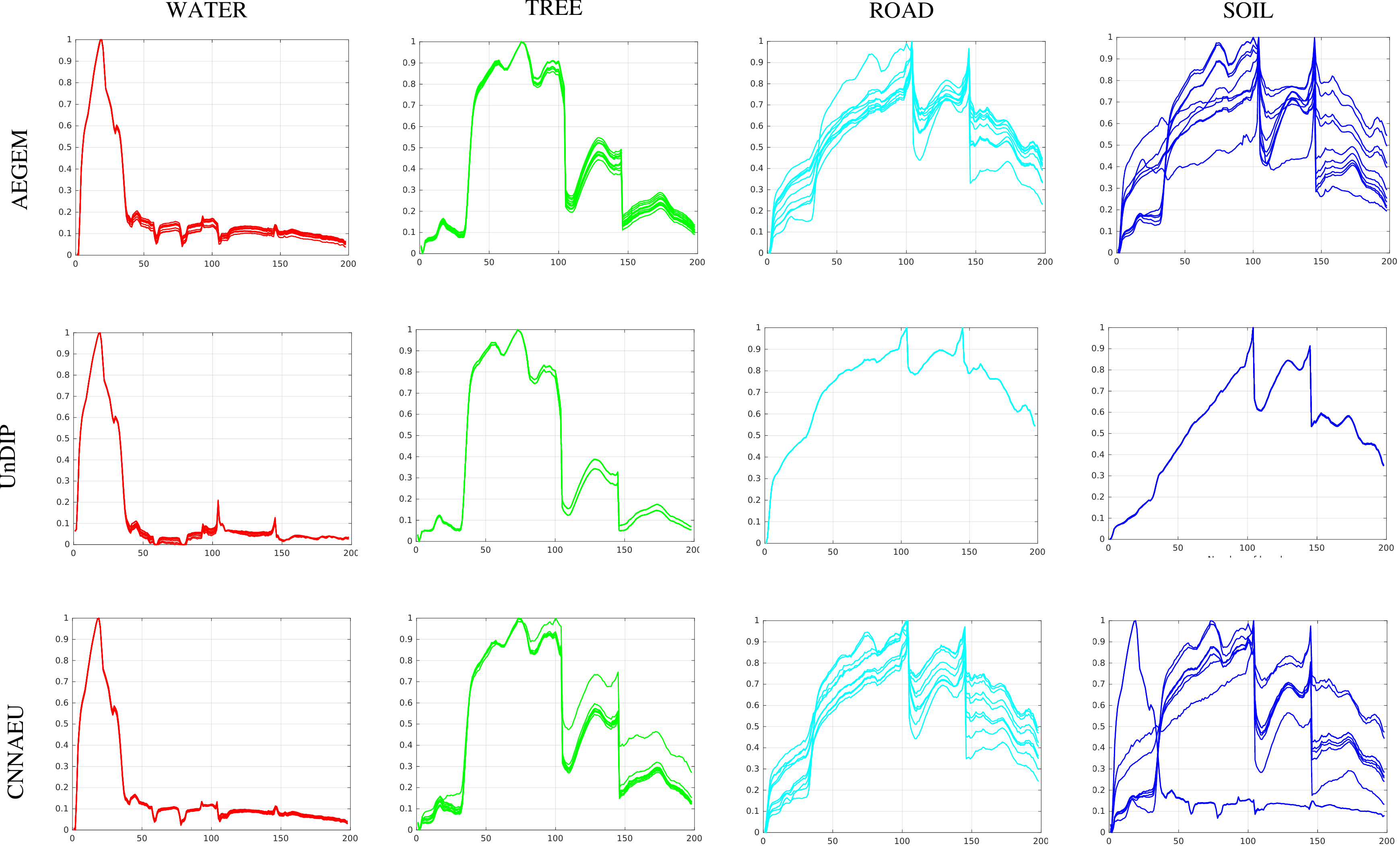}
    \caption{A comparative analysis of endmember extraction for the Jasper Ridge dataset illustrates water, tree, road, and soil extraction performed by CNNAEU, UnDIP, and the model AEGEM, conducted over 10 runs.}
\end{figure*}


\begin{table}[!h]
\renewcommand{\arraystretch}{1.3}
\caption{Comparison of the CNNAEU \cite{Palsson2021}, UnDIP \cite{Rasti2022}, and SGSNMF \cite{Yu2022_ml} with the proposed AEGEM for measuring endmember precision using the spectral angle distance and for comparing abundance maps utilizing the root mean square error on the Samson dataset.}
\label{samson_description}
\centering
\begin{tabular}{|c||c||c||c||c|}
\hline
\textbf{Material/}& \textbf{CNNAEU} & \textbf{UnDIP}& \textbf{SGSNMF}&\textbf{AEGEM}\\
\textbf{Model}& & & &\\
\hline
\textbf{Tree}&0.172&0.252&0.245&\textbf{0.158}\\
\hline
\textbf{Soil}&0.198&0.260&\textbf{0.179}&0.182\\
\hline
\textbf{Water}&0.202&0.426&0.358&\textbf{0.081}
\\
\hline
\textbf{Mean RMSE}&0.190&0.315& 0.261&\textbf{0.140}\\
\hline\hline
\textbf{Tree}&0.041&\textbf{0.022}&0.046&0.029\\
\hline
\textbf{Soil}&0.048&0.040&\textbf{0.010}&0.024\\
\hline
\textbf{Water}&0.113&0.130&0.230&\textbf{0.079}\\
\hline
\textbf{Mean SAD}&0.067&0.064&0.095&\textbf{0.044}\\
\hline
\end{tabular}
\end{table}

\begin{table}[h!]
\renewcommand{\arraystretch}{1.3}
\caption{Comparison of the CNNAEU \cite{Palsson2021}, UnDIP \cite{Rasti2022}, and SGSNMF \cite{Yu2022_ml} with the proposed AEGEM for measuring endmember precision using the spectral angle distance metric and for abundance maps utilizing the root mean square error for the Jasper dataset.}
\label{jasper_description}
\centering
\begin{tabular}{|c||c||c||c||c|}
\hline
\textbf{Material/}& \textbf{CNNAEU} & \textbf{UnDIP}&\textbf{SGSNMF}&\textbf{AEGEM}\\
\textbf{Model}& & & &\\
\hline
\textbf{Tree}&0.199&0.160&\textbf{0.140}&0.203\\
\hline
\textbf{Soil}&0.294&0.132&\textbf{0.124}&0.280\\
\hline
\textbf{Water}&0.183&0.201&0.176&\textbf{0.110}
\\
\hline
\textbf{Road}&0.308&\textbf{0.109}&0.119&0.195
\\
\hline
\textbf{Mean RMSE}&0.246&0.150&\textbf{0.140}&0.197\\
\hline
\hline
\textbf{Tree}&0.060&0.149&0.146&\textbf{0.035}\\
\hline
\textbf{Soil}&0.140&\textbf{0.114}&0.143&0.187\\
\hline
\textbf{Water}&0.061&0.252&0.226&\textbf{0.060}\\
\hline
\textbf{Road}&0.134&0.086&\textbf{0.041}&0.156\\
\hline
\textbf{Mean SAD}&0.099&0.150&0.139&\textbf{0.109}\\
\hline
\end{tabular}
\end{table}

\begin{table}[!h]
\renewcommand{\arraystretch}{1.3}
\caption{Comparison of the CNNAEU \cite{Palsson2021}, UnDIP \cite{Rasti2022}, and SGSNMF \cite{Yu2022_ml} with the proposed AEGEM for measuring endmember precision using the spectral angle metric and for the abundances maps utilizing the root mean square error on the Urban dataset.}
\label{urban_description}
\centering
\begin{tabular}{|c||c||c||c||c|}
\hline
\textbf{Material/}& \textbf{CNNAEU} & \textbf{UnDIP}&\textbf{SGSNMF} &\textbf{AEGEM}\\
\textbf{Model}& & & & \\
\hline
\textbf{Tree}&\textbf{0.127}&0.254&0.201&0.186\\
\hline
\textbf{Soil}&\textbf{0.228}&0.156&0.301&0.270\\
\hline
\textbf{Grass}&\textbf{0.240}&0.184&0.323&0.271
\\
\hline
\textbf{Roof}&0.145& 0.280&0.168&\textbf{0.135}
\\
\hline
\textbf{Asphalt}&0.245& 0.516&0.263&\textbf{0.240}
\\
\hline
\textbf{Mean RMSE}&\textbf{0.212}& 0.278&0.251&0.221\\
\hline
\hline
\textbf{Tree}&\textbf{0.087}&0.108&0.201&0.098\\
\hline
\textbf{Soil}&\textbf{0.109}&0.512&0.301&0.342\\
\hline
\textbf{Grass}&0.077&0.127&0.323&\textbf{0.063}\\
\hline
\textbf{Roof}&0.101&0.139&0.168&\textbf{0.094}\\
\hline
\textbf{Asphalt}&\textbf{0.110}&0.360&0.263&0.157\\
\hline
\textbf{Mean SAD}&\textbf{0.096}&0.242&0.251&0.150\\ 
\hline
\end{tabular}
\end{table}

\section{Conclusion}
A semi-supervised deep learning ensemble model is presented for endmembers extraction and the abundance maps computation. The AEGEM surpasses the results achieved by the state-of-the-art SU models on benchmark datasets such as Samson, Jasper, and Urban, especially for endmembers corresponding to water, soil, and asphalt.

The model AEGEM can be utilized to study various material compositions based on the analysis of spectral signatures. Due to its superior performance over water bodies, it can be employed for applications such as detecting coral reef degradation and change detection across different frames obtained at different time points. This allows for the extraction of materials and the identification of differences in each composition. 

In addition, the model AEGEM can be utilized as a classification model that exploits both spatial and spectral properties within the SU framework. This enables the identification of large spatial regions, such as land cover types, water bodies, and variations in vegetation.

\section{Conflict of interest statement}
The authors declare that the research is conducted in the absence of any commercial or financial relationships that could be construed as a potential conflict of interest.

\section{Authors contribution}

EAM, CJD, VM: conceptualization, EAM, CJD, VM: methodology, EAM, CJD: software, EAM, CJD: validation, EAM: formal analysis, EAM, CJD, VM: investigation, EAM: data curation, EAM: writing—original draft preparation, EAM, CJD, VM : writing—review and editing, VM: project administration, VM: funding acquisition; All authors contributed to the article and approved the submitted version.

\section{Data availability statement}
The Samson, Jasper Ridge, and Urban datasets are available at: https://rslab.ut.ac.ir/data (accessed on 25 November 2023).

\section{Funding}

This research is partially funded by the Artificial Intelligence Imaging Group (AIIG), Department of Electrical and Computer Engineering, University of Puerto Rico, Mayaguez. The APC is funded by NASA, grant number 80NSSC19M0155. The APC is funded by 80NSSC21M0155. Opinions, findings, conclusions, or recommendations expressed in this material are those of the authors and do not necessarily reflect the views of NASA.

\section*{Acknowledgment}

The authors would like to thank the Artificial Intelligence Imaging Group (AIIG) Laboratory and the Electrical Engineering Department at the University of Puerto Rico, Mayagüez for supporting this research work.

\ifCLASSOPTIONcaptionsoff
  \newpage
\fi

\bibliographystyle{ieeetr}
\bibliography{References}
\end{document}